\documentclass{article}

\usepackage{PRIMEarxiv}

\usepackage[utf8]{inputenc} 
\usepackage[T1]{fontenc}    
\usepackage{hyperref}       
\usepackage{url}            
\usepackage{booktabs}       
\usepackage{amsfonts}       
\usepackage{nicefrac}       
\usepackage{microtype}      
\usepackage{lipsum}
\usepackage{tablefootnote}
\usepackage{color, colortbl}
\usepackage[misc]{ifsym} 
\usepackage{soul}
\usepackage{amsmath}
\usepackage{fixltx2e}
\usepackage{afterpage}
\usepackage{caption}
\usepackage{fancyhdr}       
\usepackage{graphicx}       
\graphicspath{{media/}}     

\definecolor{brown}{RGB}{229,225,224}

\usepackage{cite}
\usepackage{graphicx}
\usepackage[flushleft]{threeparttable}

\title{Improving image quality of sparse-view lung tumor CT images with U-Net
}

\author{
  Annika Ries$^{* 1, 2}$, \Letter \hspace{0.3em} Tina Dorosti$^{* 1-3}$, Johannes Thalhammer$^\mathrm{{1-4}}$, Daniel Sasse$^\mathrm{{3}}$, Andreas Sauter$^\mathrm{{3}}$,\\
  \textbf{Felix Meurer$^\mathrm{{3}}$, Ashley Benne$^\mathrm{{3, 4}}$, Tobias Lasser$^\mathrm{{2, 5}}$, Franz Pfeiffer$^\mathrm{{1-4}}$},\\
  \textbf{Florian Schaff$^\mathrm{{1, 2}}$, Daniela Pfeiffer$^\mathrm{{3, 4}}$} \\\\
  1 Chair of Biomedical Physics, Department of Physics, School of Natural Sciences\\
  2 Munich Institute of Biomedical Engineering\\
  3 Department of Diagnostic and Interventional Radiology, School of Medicine, Klinikum rechts der Isar\\
  4 Institute for Advanced Study\\
  5 Computational Imaging and Inverse Problems, Department of Computer Science, School of Computation,\\ Information, and Technology\\\\
  Technical University of Munich, Germany\\
  * Authors contributed equally to this work\\
  \Letter \hspace{0.3em} \texttt{tina.dorosti@tum.de} \\
}

\begin{document}
\maketitle

\section*{Key points:}
\begin{enumerate}
  \item Sparse-projection-view streak artifacts reduce the quality and usability of sparse-view CT images.
  \item U-Net-based postprocessing removes sparse-view artifacts while maintaining diagnostically accurate IQ.
  \item Postprocessed sparse-view CTs drastically increase radiologists’ confidence in diagnosing lung nodules.\\
\end{enumerate}

\textbf{Abbreviations:} Computed tomography (CT), Dice similarity coefficient (DSC), Image quality (IQ), Mean squared error (MSE), Structural similarity index measure (SSIM)


\newpage
\section*{\center Declarations}
\textbf{Ethical approval and consent to participate}\\
This retrospective study obtained approval from our institutional review board and was conducted in accordance with the regulations of our institution (approval code: 87/18 S, Institutional Review Board of the Faculty of Medicine, Technical University of Munich, Germany).\\\\
\textbf{Consent for publication}\\
Not applicable.\\\\
\textbf{Availability of data and materials}\\
The CT data and the model for inference are available upon reasonable request from the corresponding author.\\\\
\textbf{Competing interests}\\
We report no conflict of interest.\\\\
\textbf{Funding}\\
Funded by the Federal Ministry of Education and Research (BMBF) and the Free State of Bavaria under the Excellence Strategy of the Federal Government and the Länder, the German Research Foundation (GRK2274), as well as by the Technical University of Munich–Institute for Advanced Study.\\\\
\textbf{Authors’ contributions}\\
All the authors contributed at the different stages of the study; AR and TD shared first authorship as they contributed equally to this work, with AR implementing the deep learning model and TD conducting the reader study at our clinic. The literature review was carried out by AR, and the statistical analysis by TD. AR, TD, JT, and DP curated and analyzed the data. DS, AS, FM, AB, and DP contributed to the clinical methodology and the reader study. AR, TD, JT, AB, TL, FP, FS, and DP investigated and interpreted the results. FP and DP conceptualized, administered, and acquired funding for this study. All authors agreed to guarantee that any questions related to this work are appropriately investigated. All authors reviewed the manuscript. All authors read and approved the final manuscript.\\\\
\textbf{Acknowledgments}\\
The authors thank Dr. Bernhard Haller for his valuable advice on the appropriate statistical methods for this work. We declare no use of large language models in our manuscript.\\\\

\newpage
\begin{abstract}
\textbf{Background:} We aimed at improving image quality (IQ) of sparse-view computed tomography (CT) images using a U-Net for lung nodule detection and determining the best tradeoff between number of views, IQ, and diagnostic confidence.\\\\
\textbf{Methods:} CT images from 41 subjects aged 62.8$\pm$10.6 (mean $\pm$ standard deviation), 23 men, 34 with lung nodules, 7 healthy, were retrospectively selected (2016--2018) and forward projected onto 2,048-view sinograms. Six corresponding sparse-view CT data subsets at varying levels of undersampling were reconstructed from sinograms using filtered backprojection with 16, 32, 64, 128, 256, and 512 views. A dual-frame U-Net was trained and evaluated for each subsampling level on 8,658 images from 22 diseased subjects. A representative image per scan was selected from 19 subjects (12 diseased, 7 healthy) for a single-blinded multireader study. These slices, for all levels of subsampling, with and without U-Net postprocessing, were presented to three readers. IQ and diagnostic confidence were ranked using predefined scales. Subjective nodule segmentation was evaluated using sensitivity and Dice similarity coefficient (DSC); clustered Wilcoxon signed-rank test was used.\\\\
\textbf{Results:} The 64-projection sparse-view images resulted in 0.89 sensitivity and 0.81 DSC, while their counterparts, postprocessed with the U-Net, had improved metrics (0.94 sensitivity and 0.85 DSC) (\textit{p} = 0.400). Fewer views led to insufficient IQ for diagnosis. For increased views, no substantial discrepancies were noted between sparse-view and postprocessed images.\\\\
\textbf{Conclusion:} Projection views can be reduced from 2,048 to 64 while maintaining IQ and the confidence of the radiologists on a satisfactory level.\\\\
\textbf{Relevance statement:} Our reader study demonstrates the benefit of U-Net postprocessing for regular CT screenings of patients with metastatic lung nodules to increase the IQ and diagnostic confidence while reducing the dose.\\\\
\textbf{Trial registration:} Not applicable\\\\


\end{abstract}
\section*{Introduction}
Lung cancer maintains the highest mortality rate for malignancies around the globe, with more than 2.2 million new cases recorded worldwide in 2020 \cite{WHO, WCRF}. More than half of all cancerous lung tumor diagnoses present as symptomatic once the patient has reached a progressive stage \cite{GEKID}. Regular screenings enable early detection and thereby increase survival rates \cite{GEKID, ACS}. \\

Computed tomography (CT) is considered standard practice in present-day medicine for diagnosing lung nodules  \cite{ACS, ONKO, NHS}, yet it comes at the cost of radiation exposure  \cite{Hamada2014, FDA}. To make regular screenings possible, a tradeoff between dose and image quality (IQ) must be found \cite{ACS}. Sparse-view CT is a technique for dose reduction. However, this technique leads to a degradation of image quality due to distinct streak artifacts caused by a limited number of projection views in the reconstruction process \cite{Kudo2013, Zhang2018}. \\

Machine learning approaches have shown promising results for sparse-view artifact correction \cite{Kudo2013, Zhang2018, Jin2016, Han2016, Han2018, Koetzier2023}. Specifically, residual learning has delivered superior results compared to the direct approach \cite{Jin2016, Han2016}. The goal of the network in residual learning is to estimate the difference between sparse-view and full-view images. In a direct approach, the network aims to predict the artifact-free image. The simpler topological structure of residual images allows for more efficient learning \cite{Han2016}. A popular network architecture for such artifact-correction tasks is the U-Net \cite{Ronneberger2015}. With a large receptive field, the model is capable of handling global artifacts such as the given sparse-view streak artifacts  \cite{Jin2016, Han2016}. The dual-frame U-Net was proposed as a more robust variant of the standard U-Net for the task at hand \cite{Han2018}. \\

In this work, we assess the performance of the dual-frame U-Net in correcting for streak artifacts present in sparse-view CT scans of the lung with metastasis \cite{Han2018}. An image reconstructed from 2,048 views, later referred to as a full-view image, was taken to calculate the residual image. Six levels of subsampled input images were reconstructed from 16, 32, 64, 128, 256, and 512 views, respectively. By conducting a reader study with the unprocessed sparse-view images and their U-Net postprocessed counterpart images, we aim to find the best tradeoff between the number of views, IQ, and confidence of the participating radiologists on their diagnosis.

\section*{Methods}
The code is available at: \\
\url{https://github.com/tidorosti/Reader-Study\_UNet-Processed-Lung-Cancer-CT} 
\subsection*{Dataset}
We received approval from the institutional review board, and the requirement for written informed consent was waived, as all original data (acquired between January 2016 and December 2018) was analyzed anonymously and retrospectively. Seven CT images from seven subjects without lung metastasis, additional pleural effusion, atelectasis, or other lung diseases were selected as the healthy controls. A total of 16,023 CT images from 42 subjects were considered for the diseased group such that all images presented exactly one lung metastatic nodule of size roughly from 1 to 2 cm in diameter. As we aimed to focus solely on subjects with lung metastasis without other lung diseases, the following exclusion process was applied: after the elimination of cases with perihilar localization of metastases, 14,578 images from 38 subjects remained. Next, images with additional pleural effusion, atelectasis, or other lung diseases were removed. Finally, 8,670 images from 34 subjects with metastatic lung nodules were selected as the diseased group. The complete dataset consisted of 8,677 images from 41 subjects (34 diseased and 7 healthy subjects). From this, independent datasets were utilized for model assessment (8,658 images from 22 diseased subjects) and the reader study (19 images corresponding to one image per subject; 12 diseased and 7 healthy subjects).  Additional 9,481 images from the Luna16 external dataset \cite{vanGinneken2016, Armato2011} were utilized for testing the model’s robustness. Table \autoref{tab:demographics} shows the subject demographics for the internal datasets.

\begin{table}[h!]
\centering
\begin{threeparttable}
\caption{\small Subject demographics for internal datasets  (\textit{n} = 41)}
\label{tab:demographics}
\small
  \centering
    \begin{tabular}{lcccccc}
    \toprule 
    Dataset & \multicolumn{3}{c}{Model assessment} & \multicolumn{2}{c}{Reader study} \\
    \midrule
    \rowcolor{brown}Parameter \textbackslash    subset &  Train  &  Validation & Test &  Healthy  &  Diseased \\
    \midrule
    \midrule
    Male & 5 & 2 & 4 & 5 & 7\\
    \rowcolor{brown}Female & 7 & 0 & 4 & 2 & 5\\
    Age (years) & 65.8 $\pm$ 11.6 & 77.0 $\pm$ 4.00 & 61.9 $\pm$ 11.8 & 44.3 $\pm$ 14.4 & 64.8 $\pm$ 9.25\\
    \rowcolor{brown}Total CT Images & 4,723 & 787 & 3,148 & 7 & 12\\
    \bottomrule
    \end{tabular}
    \begin{tablenotes}
    \small
      \item Age is given as mean $\pm$ standard deviation.
    \end{tablenotes}
    \end{threeparttable}
\end{table}

\subsection*{Data Preparation}
The CT images were forward projected onto 2,048-view sinograms. Sparse-view CT data subsets at varying levels of undersampling were generated using the filtered back projection algorithm with 16, 32, 64, 128, 256, and 512 views, respectively. The full-view data was generated using 2,048 views. All operations were performed using the Astra toolbox (version 2.1.1) \cite{vanAarle2015, vanAarle2016, Palenstijn2011}. Images were of size 512 $\times$ 512 pixels. The intensity values of all images were clipped to the lung CT window (width 1,700, level -600 HU) and normalized to a range between zero and one. \\

Twenty-two of the diseased subjects were split on CT scan level into train (\textit{n} = 12, images = 4,723), validation (\textit{n} = 2, images = 787), and test sets (\textit{n} = 8, images = 3,148). The residual ground truth label images were calculated as the difference between the full-view and the sparse-view images for each projection view. The final postprocessed image was the pure-artifact U-Net prediction subtracted from the sparse-view input.  

\subsection*{Network Architecture}
\begin{figure*}[t]
\centerline{\includegraphics[width=\textwidth, trim={0 0.75cm 0 0}, clip]{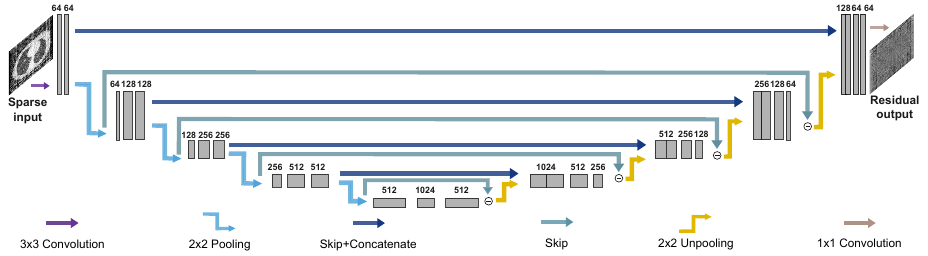}}
\caption{\small The architecture of the dual-frame U-Net. The model takes as input the unprocessed sparse-view images and outputs the pure artifact residual image. An example of 16 projection sparse-view input and corresponding residual output is shown. The number of channels is provided above each layer.}
\label{fig:dualUnet}
\end{figure*}

The dual-frame U-Net was utilized, as depicted in \autoref{fig:dualUnet}. The contracting path consists of four subsequently applied encoder blocks, each with two convolution layers (3 $\times$ 3 kernels, followed by batch normalization and a rectified linear unit activation). A 2 $\times$ 2 max pooling layer is applied after each encoder block. Following the two convolution layers in the bottleneck, the features are upsampled with four subsequently applied decoder blocks mirroring the contracting path via a 2 $\times$ 2 upsampling with nearest neighbor interpolative resizing before each decoder block. The dual-frame U-Net introduces additional skip connections, bridging the output of each encoder block after pooling to the input of the associated decoder block before upsampling. These additional connections ensure the frame condition is met, thereby reducing blurring and image artifacts. The final image is obtained with a 1 $\times$ 1 convolution \cite{Han2018}

A train-validation-test split method was chosen instead of a cross-validation method due to time and computation constraints. The training data was randomly selected from all the available internal data on a patient level using Python’s built-in random function. The proposed model was additionally tested on an external test set to ensure the robustness of the final model, and it was concluded that the train-validation-test split method did not hinder model performance. \\

An NVIDIA RTX A4000 graphic card with 16 GB of VRAM was utilized to train this dual-frame U-Net with 21,971,584 parameters. The network was implemented by the Keras interface of the TensorFlow library (version 2.4.0), randomly initialized, and trained individually for each number of projection views \cite{Keras, Tensorflow}. The sparse-view images were taken as input, and the residual images were taken as labels. No data augmentation was applied as the model achieved comparable results for the training and validation set without overfitting. Mean squared error (MSE) loss with an adaptive moment estimation optimizer was utilized. Early stopping was implemented if validation loss did not improve. Training took place for a maximum of \textit{n} = 30
epochs and a batch size of six. The initial learning rate
$lr$ was set to $lr = 0.001$ and decayed exponentially per epoch following $lr_n = lr_{n-1} \cdot e^{-0.1}$ . The
model with the smallest validation loss among all epochs was chosen for inference on the test sets
and the reader study. The quality of postprocessed images was evaluated with the MSE and the structural similarity index measure (SSIM) metrics \cite{SSIM}.\\

The dual-frame U-Net was chosen as it generated robust outputs and had a comparable computational effort as the standard U-Net. More specifically, The test data was analyzed with both the dual-frame and the standard U-Net, and there were no major differences in the MSE and SSIM values between the two models. Furthermore, the number of model parameters and the computation time were also comparable. Lastly, our expert radiologist (D.P.) examined the data and concluded that images postprocessed with the dual-frame U-Net more accurately display medically relevant structures, such as small vessels.

\subsection*{Multireader study and statistical analysis}
CT scans from 19 subjects (12 diseased, 7 healthy) were considered for this single-blinded study. Three board-certified radiologists and an in-training radiologist, respectively with 15 (D.P.), 11 (A.S.), 10 (F.M.), and 5 (D.S.) years of experience in chest radiology, participated in the study. Using the full-view images, D.P. selected a representative slice per subject and marked the ground truth lung nodule segmentation ($1.11 [0.91, 1.31] cm$ diameter given as mean with 95\% confidence interval) for the diseased subjects. All nodules were confirmed metastases by biopsy, patient history, and follow-up procedures. The sparse-view images reconstructed from 16, 32, 64, 128, and 256 views and postprocessed by the U-Net were presented to the other three radiologists, resulting in a total of 190 evaluated images per reader.\\

\begin{figure*}[t]
\centerline{\includegraphics[width=\textwidth]{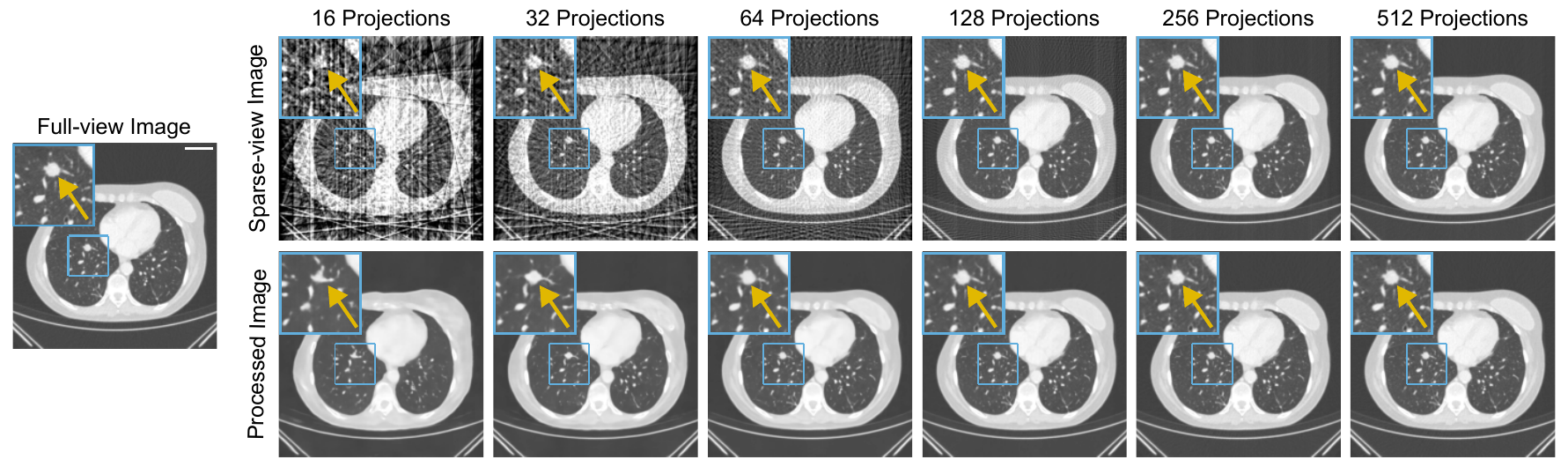}}
\caption{\small An example computed tomography (CT) image reconstructed with full-view and sparse-view projections, with and without postprocessing by the dual-frame U-Net. The image on the left demonstrates the ground truth full-view image without postprocessing. The top row shows the CT image reconstructed with different sparse-view projections without postprocessing. The bottom row depicts the respective sparse-view images postprocessed by the U-Net model for each projection view. The region of interest (blue box) shows the metastasis (highlighted by the yellow arrow). All images are clipped to the lung window and include an iodined contrast medium. Scale bar in the full-view image = $5 cm$ .}
\label{fig:exampleSlices}
\end{figure*}

Full-view and all sparse-view images of an exemplary slice are shown in \autoref{fig:exampleSlices}. Slices reconstructed and postprocessed using 512 views were excluded from the study as D.P. determined that even without any postprocessing, they are of comparable quality to the full-view images.\\

Readers were asked to independently annotate each slice using our in-house tool by rating every image on quality, the confidence of diagnosis, and the severity of artifacts present in the image according to pre-defined labels in Tables \autoref{tab:labels} and \autoref{tab:labels_artifact}. Furthermore, the radiologists were asked to independently segment perceived suspect pulmonary nodules. Sensitivity, specificity, F\textsubscript{1} score, and the negative predictive value, were considered to compare the diagnostic reliability of images for different views \cite{Parikh2008}. For all true positive cases, the segmentation overlaps were calculated with the Dice similarity coefficient (DSC) \cite{Dice1945, Fleiss2003}. In case of no overlap, or if one of the segmentations was empty, the resulting DSC was zero. \\

The superiority of the postprocessed labeled data over the sparse-view labeled data for each view was assessed: p-values were calculated with the clustered Wilcoxon signed-rank test utilizing Python’s SciPy library (version 1.4.1), and a significance threshold of 0.05 was set \cite{Wilcoxon, Virtanen2020}. The sample size for the reader study was \textit{n} = 57 after pooling the results from the three readers, with each having annotated 19 CT images. \\

\begin{table}[h!]
\caption{\small Score system for image quality and diagnostic confidence}
\label{tab:labels}
\small
  \centering
    \begin{tabular}{lll}
    \toprule 
    Scale & Quality & Confidence \\
    \midrule
    \midrule
    \rowcolor{brown} 1 & Not diagnostic & Not confident at all \\
    2 & Highly impaired & Slightly confident \\
    \rowcolor{brown} 3 & Impaired & Somewhat confident \\
    4  & Sufficient & Fairly confident \\
    \rowcolor{brown} 5 & High & Very confident \\
    6 & Very high & Surely confident\\
    \bottomrule
    \end{tabular}
\end{table}

\begin{table}[h!]
\caption{\small Score system for image artifacts }
\label{tab:labels_artifact}
\small
  \centering
    \begin{tabular}{ll}
    \toprule 
    Scale& Artifacts \\
    \midrule
    \midrule
    \rowcolor{brown} 1 & No artifacts \\
    2  & Few artifacts and quality not impaired\\
    \rowcolor{brown} 3 & Some artifacts and reduced quality\\
    4  & A lot of artifacts and reduced quality \\
    \bottomrule
    \end{tabular}
\end{table}

\section*{Results}
The following results show the model's performance on 3,148 images from eight diseased subjects and 9,481 images from the Luna16 dataset. Furthermore, results of the reader study on 19 CT-wise images from 12 diseased and seven healthy subjects are described.

\subsection*{Network Performance}

\autoref{fig:exampleSlices} shows an example slice with varying levels of subsampling alongside the corresponding U-Net postprocessed results. It can be observed that fewer projection views result in more artifacts. The sparse-view images from extremely limited views also lead to a loss of structural integrity in their postprocessed counterparts. This was especially prominent for 16 views, as metastatic lung nodule distortion and microvascular structures generate diminished performance capabilities. metastatic nodule composition and primary anatomical characteristics can better be amassed once reconstruction views have increased to 32. For 64 views, streak artifacts did not impact the nodule’s visibility due to tissue density, but minimalistic structural identification, such as small vessels, are not clearly portrayed. Minor features were displayed for 128 and 256 views; however, for 128 views, some streak artifacts remained present. For the postprocessed image of 32 views, the nodule shape was mostly correct, and the display of the vascular structures was improved. For 64 or more views, the nodule appearance in the postprocessed image was similar to the full-view image. Furthermore, vascular distinction on imaging can be detected with the postprocessed 128-view image. The postprocessed image from 256 views is very close in quality to the full-view image. For 512 views, no qualitative differences can be detected.\\

A directly proportional relationship is observed between improved IQ and higher views. As shown in \autoref{tab:MSE_SSIIM}, calculated mean MSE values decrease and mean SSIM values increase with more projection views for the internal test set and the external Luna16 dataset. Although mean MSE and SSIM values are marginally better for the internal test set, the model achieves comparable results on the external Luna16 dataset. 

\begin{table*}[h!]
\centering
\begin{threeparttable}
\captionof{table}{\small Mean MSE and SSIM \\}
\label{tab:MSE_SSIIM}
  \small

   \begin{tabular}{l|l|ccc}
    \toprule
     Metric & Dataset & 16 projections& 32 projections& 64 projections \\
    \midrule
    \midrule
    \rowcolor{brown} MSE & Test set & $2.36 [1.95, 2.78] \cdot 10^{-3}$ & $7.95 [6.51, 9.39] \cdot 10^{-4}$ &$2.40 [1.96, 2.84] \cdot 10^{-4}$ \\
    & Luna16 & $6.52 [4.93, 8.10] \cdot 10^{-3}$ & $3.46 [2.49, 4.43] \cdot 10^{-3}$ &$1.04 [0.746, 1.34] \cdot 10^{-3}$ \\
    \rowcolor{brown} SSIM & Test set & $0.799 [0.749, 0.809]$ & $0.834 [0.808, 0.861]$ & $0.895 [0.873, 0.917]$ \\
    & Luna16 & $0.782 [0.758, 0.805]$ & $0.816 [0.792, 0.840]$ & $0.873 [0.852, 0.895]$ \\
    \midrule
    \midrule
     &  & 128 projections& 256 projections & 512 projections \\
    \rowcolor{brown} MSE & Test set & $8.46 [6.31, 10.6] \cdot 10^{-5}$ & $2.28 [1.54, 3.01] \cdot 10^{-5}$ &$3.78 [2.81, 4.75] \cdot 10^{-6}$ \\
    & Luna16 & $6.19 [4.18, 8.19] \cdot 10^{-4}$ & $1.07 [0.810, 1.32] \cdot 10^{-4}$ &$5.34 [3.93, 6.75] \cdot 10^{-5}$ \\
    \rowcolor{brown} SSIM & Test set & $0.938 [0.920, 0.955]$ & $0.979 [0.973, 0.985]$ & $0.997 [0.996, 0.997]$ \\
    & Luna16 & $0.908 [0.889, 0.927]$ & $0.960 [0.950, 0.969]$ & $0.983 [0.980, 0.986]$ \\
    \bottomrule
    \end{tabular}
    \begin{tablenotes}
      \small
      \item Postprocessed images in the internal test set and the external Luna16 dataset for all projection views are presented by the mean value and the corresponding 95\% confidence intervals of the mean squared error (MSE) and structural similarity index measure (SSIM) metrics.
    \end{tablenotes}
\end{threeparttable}
\end{table*}

\subsection*{Multireader Study}
\begin{figure*}[!]
\centerline{\includegraphics[width=\textwidth, trim={0 5.5cm 0 5.5cm},clip]{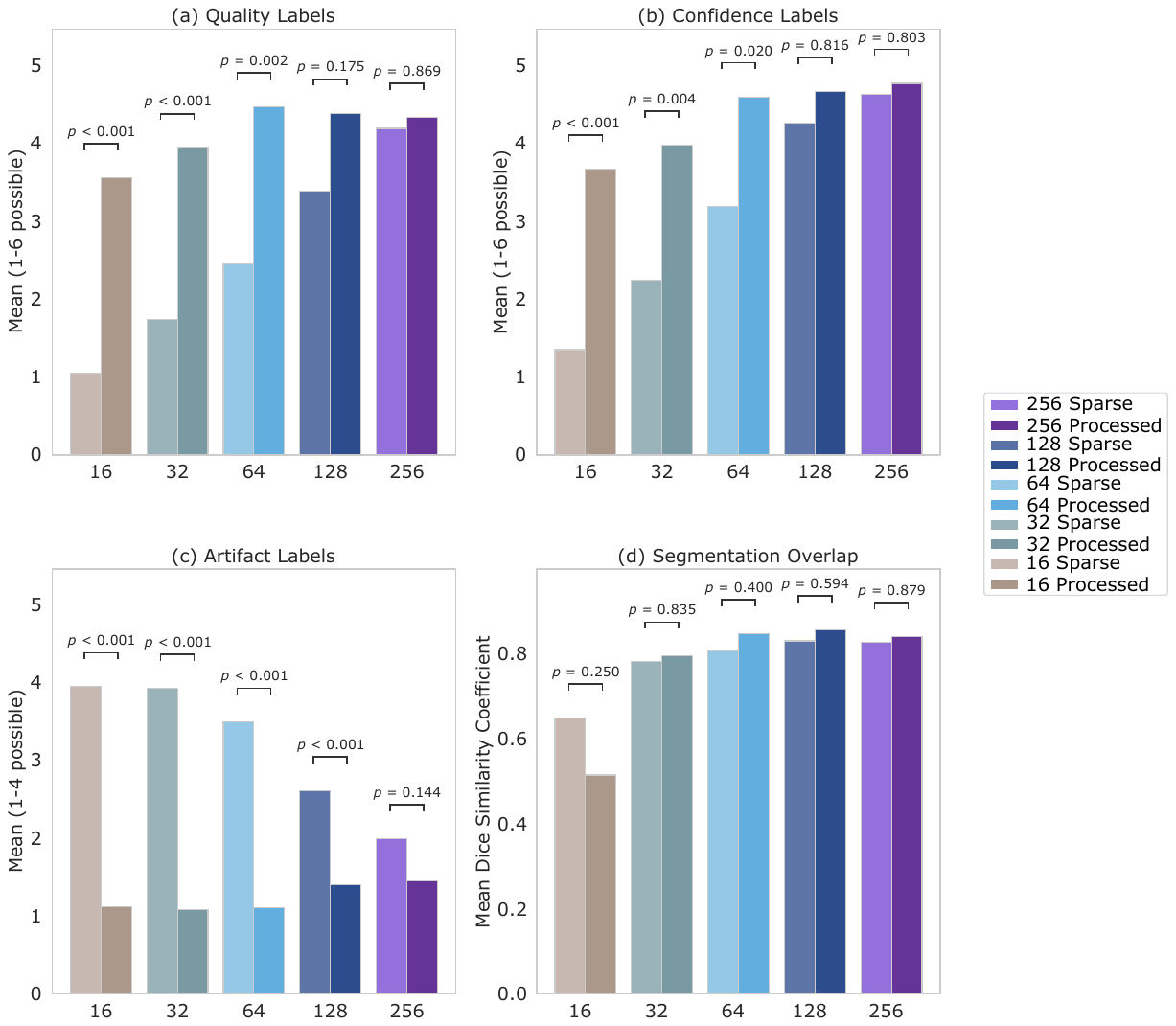}}
\captionof{figure}{\small Mean over image quality (\textbf{a}), diagnostic confidence (\textbf{b}), severity of artifacts (\textbf{c}), and Dice similarity coefficient values (\textbf{d}) for lung nodule segmentations for 19 sparse-view images with (processed) and without postprocessing (sparse) by the dual-frame U-Net, labeled by three readers (\textit{n} = 57). Scales defined for all labels are given in Tables \autoref{tab:labels} and \autoref{tab:labels_artifact}.}\label{fig:labels}
\end{figure*}

The resulting mean values for quality, confidence, and artifacts reported by the readers are shown in \autoref{fig:labels}a-c. The labeled mean quality for sparse-view images decreases linearly from roughly “sufficient” to approximately “not diagnostic” for decreasing number of projection views, as seen in \autoref{fig:labels}a. \autoref{fig:labels}b shows that the tendency for the mean confidence is similar for both sparse-view and postprocessed images. For the sparse-view images, the confidence again decreases linearly with decreasing number of views, ranging from “fairly confident” or “very confident” to “not confident at all” or “slightly confident.” The subjective quality (\textit{p} = 0.002) and confidence (\textit{p} = 0.020) of postprocessed images are significantly higher than their unprocessed pairs for 64 and fewer views. The presence of artifacts increases for the sparse-view images with fewer views, as observed in \autoref{fig:labels}c. Postprocessed images have significantly fewer subjective artifacts than their unprocessed pairs for 128 and fewer projection views (\textit{p} < 0.001).\\

\begin{figure*}[!]
\centerline{\includegraphics[width=\textwidth, trim={0 10cm 0 10cm},clip]{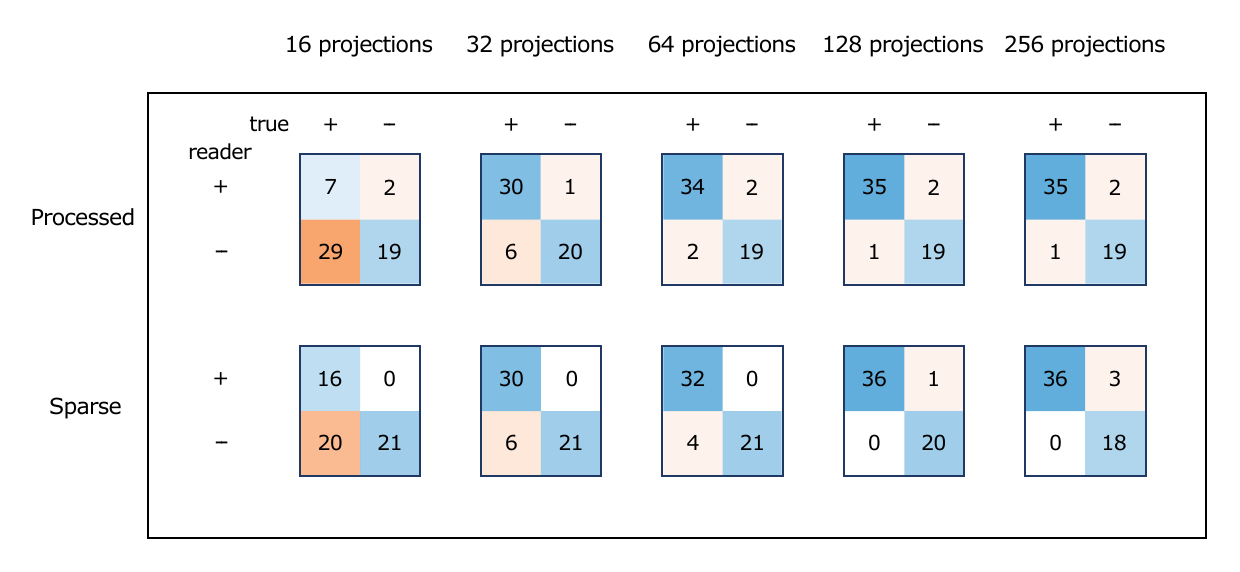}}
\captionof{figure}{\small Confusion matrices for sparse-view CT images and their postprocessed counterpart images for all projection views were calculated over 19 subject-wise images presented to three readers (\textit{n} = 57).}\label{fig:CMs}
\end{figure*}

\begin{figure*}[t]
\centerline{\includegraphics[width=\textwidth]{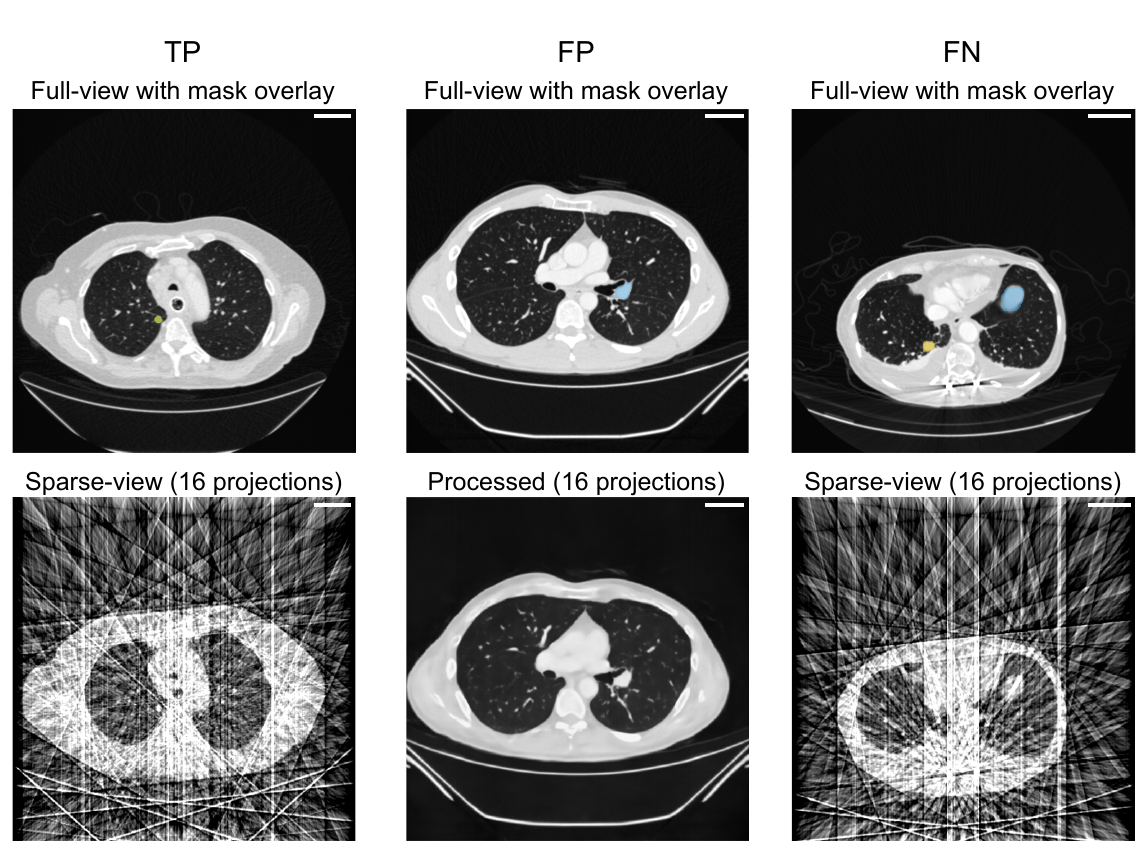}}
\captionof{figure}{\small Examples of metastasis segmentations. A correctly marked nodule, true positive (TP), and two incorrectly segmented regions, namely false negative (FN) and false positive (FP), are shown. FP refers to the case where the perceived metastasis was nonexistent. FN refers to the case where the perceived nodule had no overlap with the ground truth segmentation. The top row shows the overlay of the ground truth segmentation (yellow) and the segmentation marked by the reader (blue) over the full-view image. The bottom row shows the sparse-view image, reconstructed from 16 projection views with or without postprocessing, presented to the readers for marking lung nodules. All slices are clipped to the lung window and include an iodined contrast medium. Scale bar
 = $ 5 cm$.}
\label{fig:failureAnalysis}
\end{figure*}

Confusion matrices are shown in \autoref{fig:CMs}. The corresponding sensitivity, specificity, F\textsubscript{1} score, and negative predictive values are shown in Table \autoref{tab:SeSp}. In some images, incorrect subjective segmentation by the readers resulted in falsely marked pixels in an alternate location. Such cases are counted as false negatives and mostly appeared for the sparse-view images reconstructed with 16 views. An example of such an inaccurately marked image, as well as a correctly marked image, and an image with an extra perceived nodule, are shown in \autoref{fig:failureAnalysis}.\\

The confusion matrices in  \autoref{fig:CMs} show increasing false negative cases with a decreasing number of views for the sparse-view images and their postprocessed counterparts. This leads to a decreased sensitivity, as seen in Table \autoref{tab:SeSp}. The symmetric representation of true positive rate and sensitivity is understood with the F\textsubscript{1} score: For 256 and 64 views, the F\textsubscript{1} score remains unchanged among the sparse-view and the postprocessed pairs. For all other projection views, the F\textsubscript{1} score is higher for the sparse-view images. Furthermore, the number of false positive cases is mostly independent of the number of views, which leads to specificity values between 0.86 and 1.00. The negative predictive value decreases with decreasing projection views for both sparse-view and postprocessed images. However, only for 64 views do the postprocessed images achieve a higher negative predictive value compared to their sparse-view counterparts. \\

\autoref{fig:labels}c shows the mean DSC for sparse-view images with and without postprocessing by the model. The mean DSC shows only slight differences between sparse-view images with and without postprocessing for 32 or more views. For instance, in the case of 64 views, sparse-view images without postprocessing resulted in DSC = 0.81, while images postprocessed by the model had reached DSC = 0.85 (\textit{p} = 0.400). It must be noted that although no statistically significant discrepancy in segmentation overlap is observed, subjective quality (\textit{p} = 0.002) and confidence (\textit{p} = 0.020) assessment was markedly higher in the postprocessed images of 64 views and fewer.

\begin{table*}[h!]
\captionof{table}{\small Sensitivity, specificity, F\textsubscript{1} score, and negative predictive value (NPV) for sparse-view CT images and their postprocessed counterpart images for all projection views calculated over 19 subject-wise images presented to three readers (\textit{n} = 57)\\}
\label{tab:SeSp}
  \small
  \centering
   \begin{tabular}{l|l|ccccc}
    \toprule
     \multicolumn{1}{c}{}& \multicolumn{1}{c}{}& 16 & 32 & 64 & 128 & 256 \\
     \multicolumn{1}{c}{}&  \multicolumn{1}{c}{}& projections & projections & projections & projections & projections\\
    \midrule
    \midrule
    \rowcolor{brown} Processed & Sensitivity & 0.19 & 0.83 & 0.94 & 0.97 & 0.97\\
    & Specificity & 0.90 & 0.95 & 0.90 & 0.90 & 0.90\\
    \rowcolor{brown}  & F\textsubscript{1} &0.31&	0.90&	0.94&	0.96&	0.96\\
    & NPV & 0.40&	0.77&	0.90&	0.95&	0.95\\
    \rowcolor{brown} Sparse & Sensitivity & 0.44 & 0.83 & 0.89 & 1.00 & 1.00\\
    & Specificity & 0.86 & 1.00 & 1.00 & 0.95 & 0.86\\
    \rowcolor{brown}  & F\textsubscript{1} &0.62&	0.91&	0.94&	0.99&	0.96\\
    & NPV & 0.51	&0.78	&0.84	&1.00	&1.00\\
    \bottomrule
    \end{tabular}
\end{table*}


\section*{Discussion}
We implemented a postprocessing correction with a dual-frame U-Net based on a residual approach to improve the IQ of sparse-view CT images with lung metastasis. External evaluation with a public dataset demonstrated the model robustness. Furthermore, a single-blinded reader study determined a tradeoff between the number of projection views, IQ, and diagnostic confidence. The results suggest that postprocessing by the U-Net can reduce the number of views from 2,048 to only 64 while maintaining diagnostically accurate IQ for nodule detection (sensitivity = 0.94). Although the DSC for the lung nodule segmentations by the readers did not significantly improve for the postprocessed images, the sparse-view artifact-corrected images drastically increased the readers’ confidence in detecting lung nodules.\\

It must be noted that every image labeled as “not diagnostic” in terms of IQ or “not confident at all” in terms of confidence of diagnosis would not be considered in a clinical workflow. This is especially the case for sparse-view images reconstructed from 16 views but also for some sparse-view images reconstructed from 32 views. Thus, these instances will not be considered for further discussion.\\

All images postprocessed by the model are labeled with better IQ and diagnostic confidence. More precisely, the difference between sparse-view images with and without postprocessing is the most prominent result for all assigned labels. It indicates that the radiologists prefer working with the postprocessed images over the unprocessed sparse-view ones: they rate their quality higher, see fewer artifacts in the images, and, most importantly, are more confident in their diagnosis. Especially the higher quality and the increased confidence could be accompanied by a shorter processing time and, in the long run, lead to fewer signs of fatigue compared to working with unprocessed sparse-view images. Since 256, 128, and 64 views lead to very similar results regarding the quality and confidence labels and worse results are achieved with 32 views, 64-view images appear to be the best choice.\\

To define a threshold providing a reasonable tradeoff between a reduced number of projection views and diagnostic value, sensitivity, and specificity values should be maximized. Accordingly,  false positive and false negative values should be minimized: false positive cases should be avoided as these cause unnecessary follow-up procedures, potentially exposing the patient to more radiation if a full-view scan is required. However, it is of utmost importance to avoid false negative cases since these would lead to afflicted patients not getting diagnosed. Low false positive cases are correlated with high specificity, and low false negative values are associated with high sensitivity.\\

We must consider other existing work in the literature to establish concrete baseline threshold values for sensitivity and specificity. However, finding fitting pre-defined thresholds for sensitivity and specificity values proves difficult in the extant literature. This is mainly due to the challenges of establishing a truth value from which the performance of radiologists in lung nodule detection should be assessed \cite{Armato2009}. Furthermore, the variability in study design and data are limiting factors \cite{Armato2009, Rubin2015}. Nonetheless, we take the values presented in the National Lung Screening Trial by Aberle et al. \cite{Aberle2013} as the closest established baselines to which we can compare the values obtained in our study: these are a sensitivity threshold of 0.94 and a specificity threshold of 0.73. According to these thresholds, the lowest possible number of projection views allowing reliable diagnosis would be achieved for postprocessed images of 64 views, leading to 0.94 sensitivity and 0.90 specificity. \\

The mean DSC values did not consistently show a trend of improvement between the postprocessed and the unprocessed sparse-view images. Yet, these findings support the choice of the tradeoff threshold at 64 views: the mean DSC values for the postprocessed images of 64 views result in the greatest improvement over the mean DSC values of their unprocessed counterparts in comparison to the other projection views.\\

Some study limitations must be considered. In clinical practice, radiologists often search the entire stack of images for malignancies. The present reader study could have modeled the clinical workflow more precisely as it only considered single CT images. Including neighboring slices would come closer to clinical diagnosis based on CT scans and most likely reduce the amount of falsely classified patients. Furthermore, the sparse-view data generated for this study was obtained using simplified conditions not reflective of the complex reconstruction processes in clinical settings. Therefore, only the reduced number of projection views compared to the full-view images can be reported, and an exact measure of dose reduction is hence unachievable. Our relatively small sample size was also a limiting factor, which can be addressed in future works. Additionally, testing for noninferiority or equivalence of U-Net-based postprocessing with the existing methods needs further exploration before integration of such technologies in the medical workflow.\\

Overall, the amount of projection views can be reduced by a factor of 32 compared to the full-view image with postprocessing by a dual-frame U-Net while keeping the diagnostic value and the confidence of the radiologists at a satisfactory level. Regarding the radiologists’ confidence, the images postprocessed with the model lead to drastically better results than the unprocessed sparse-view images. These findings suggest that postprocessed sparse-view CT images by the dual-frame U-Net could help enable dose-efficient screening for lung metastasis detection.



\newpage
\begin{thebibliography}{00}

\bibitem{WHO}World Health Organization (2022) Cancer. Available at: \url{https://www.who.int/news-room/fact-sheets/detail/cancer}. Accessed February 10, 2023

\bibitem{WCRF}World Cancer Research Fund International (2022) Lung cancer. Available at: \url{https://www.wcrf.org/cancer-trends/lung-cancer-statistics/}. Accessed March 20, 2023

\bibitem{GEKID}Gesellschaft der epidemiologischen Krebsregister e.V. und Zentrum für Krebsregisterdaten im Robert Koch-Institut (2018) Krebs in Deutschland. Available at: \url{https://www.krebsdaten.de/Krebs/DE/Content/Publikationen/Krebs_in_Deutschland/kid_2021/kid_2021_c33_c34_lunge.pdf?__blob=publicationFile}. Accessed February 10, 2023

\bibitem{ACS}American Cancer Society (2023) Lung cancer. Available at: \url{https://www.cancer.org/cancer/lung-cancer.html}. Accessed February 10, 2023

\bibitem{ONKO}Deutsche Krebsgesellschaft (2013) Lungenkrebs / Lungenkarzinom. Available at: \url{https://www.krebsgesellschaft.de/onko-internetportal/basis-informationen-krebs/krebsarten/lungenkrebs.html}. Accessed February 10, 2023

\bibitem{NHS}
National Health Service (2022) Overview - Lung cancer. Available at: \url{https://www.nhs.uk/conditions/lung-cancer/}. Accessed February 10, 2023
\bibitem{Hamada2014}Hamada N, Fujimichi Y (2014) Classification of radiation effects for dose limitation purposes: history, current situation and future prospects. J Radiat Res 55:629–640. \url{https://doi.org/10.1093/jrr/rru019}

\bibitem{FDA}US Food and Drug Administration, Center for Devices and Radiological Health (2018) What are the radiation risks from CT? Available at: \url{https://www.fda.gov/radiation-emitting-products/medical-x-ray-imaging/what-are-radiation-risks-ct}. Accessed February 10, 2023

\bibitem{Kudo2013}Kudo H, Suzuki T, Rashed EA (2013) Image reconstruction for sparse-view CT and interior CT-introduction to compressed sensing and differentiated backprojection. Quant Imaging Med Surg 3:147–161. \url{https://doi.org/10.3978/j.issn.2223-4292.2013.06.01}

\bibitem{Zhang2018}Zhang Z, Liang X, Dong X, Xie Y, Cao G (2018) A sparse-view CT reconstruction method based on combination of denseNet and deconvolution. IEEE Trans Med Imaging 37:1407–1417. \url{https://doi.org/10.1109/TMI.2018.2823338}

\bibitem{Jin2016}Jin KH, McCann MT, Froustey E, Unser M (2016) Deep convolutional neural network for inverse problems in imaging. IEEE Trans Image Process 26:4509–5422. \url{https://doi.org/10.1109/TIP.2017.2713099}

\bibitem{Han2016}Han Y, Yoo J, Ye JC (2016) Deep residual learning for compressed sensing CT reconstruction via persistent homology analysis. arXiv. Available at: \url{https://arxiv.org/abs/1611.06391}. Accessed February 10, 2023

\bibitem{Han2018}Han Y, Ye JC (2018) Framing U-Net via deep convolutional framelets: application to sparse-view CT. IEEE Trans Med Imaging 37:1418–1429. \url{https://doi.org/10.1109/TMI.2018.2823768}

\bibitem{Koetzier2023}Koetzier LR, Mastrodicasa D, Szczykutowicz TP et al (2023) Deep learning image reconstruction for CT: technical principles and clinical prospects. Radiology 306:e221257. \url{https://doi.org/10.1148/radiol.221257}

\bibitem{Ronneberger2015}Ronneberger O, Fischer P, Brox T (2015) U-Net: convolutional networks for biomedical image segmentation. In: Navab N, Hornegger J, Wells W, Frangi A (eds) MICCAI. 18th International conference on medical image computing and computer-assisted intervention, Munich, October 2015. Lecture notes in computer science(), vol 9351. Springer, Cham, p 234–241. \url{https://doi.org/10.1007/978-3-319-24574-4_28}

\bibitem{vanGinneken2016}van Ginneken B, Setio AAA, Jacobs C (2016) Lung nodule analysis. Available at: \url{https://luna16.grand-challenge.org}. Accessed January 16, 2020

\bibitem{Armato2011}Armato SG 3rd, McLennan G, Bidaut L et al (2011) The lung image database consortium (LIDC) and image database resource initiative (IDRI): a completed reference database of lung nodules on CT scans. Med Phys 38:915–31. \url{https://doi.org/10.1118/1.3528204}

\bibitem{vanAarle2015}van Aarle W, Palenstijn WJ, De Beenhouwer J et al (2015) The ASTRA toolbox: a platform for advanced algorithm development in electron tomography. Ultramicroscopy 157:35–47. \url{https://doi.org/10.1016/j.ultramic.2015.05.002}

\bibitem{vanAarle2016}van Aarle W, Palenstijn WJ, Cant J et al (2016) Fast and flexible X-ray tomography using the ASTRA toolbox. Opt Express 24:25129–25147. \url{https://doi.org/10.1364/OE.24.025129}
\bibitem{Palenstijn2011}Palenstijn WJ, Batenburg KJ, Sijbers J (2011) Performance improvements for iterative electron tomography reconstruction using graphics processing units (GPUs). J Struct Biol 176:250–253. \url{https://doi.org/10.1016/j.jsb.2011.07.017}

\bibitem{Keras}Chollet F et al (2015) Keras. Available at: \url{https://keras.io}. Accessed February 10, 2023

\bibitem{Tensorflow}Abadi M, Agarwal A, Barham P et al (2015) TensorFlow: large-scale machine learning on heterogeneous systems. Available at: \url{https://www.tensorflow.org/}. Accessed February 10, 2023

\bibitem{SSIM} Wang Z, Bovik AC, Sheikh HR, Simoncelli EP (2004) Image quality assessment: from error visibility to structural similarity. IEEE Trans Image Process 13:600–612. \url{https://doi.org/10.1109/TIP.2003.819861}

\bibitem{Parikh2008}Parikh R, Mathai A, Parikh S, Sekhar CG, Thomas R (2008) Understanding and using sensitivity, specificity and predictive values. Indian J Ophthalmol 56:45–50. \url{https://doi.org/10.4103/0301-4738.37595}

\bibitem{Dice1945}Dice LR (1945) Measures of the amount of ecologic association between species. Ecology 26:297–302. \url{https://doi.org/10.2307/1932409}

\bibitem{Fleiss2003}Fleiss JL, Levin B, Paik MC (2003) The measurement of interrater agreement. In: Shewart WA, Wilks SS (eds). Statistical methods for rates and proportions 2003. John Wiley \& Sons, p 598–626. \url{https://doi.org/10.1002/0471445428.ch18}

\bibitem{Wilcoxon} Wilcoxon F (1945) Individual comparisons by ranking methods. Biometrics Bulletin 1:80–83. \url{https://doi.org/10.2307/3001968}

\bibitem{Virtanen2020}Virtanen P, Gommers R, Oliphant TE et al (2020) SciPy 1.0: Fundamental Algorithms for Scientific Computing in Python. Nat Methods 17:261–272. \url{https://doi.org/10.1038/s41592-019-0686-2}

\bibitem{Armato2009}Armato SG 3rd, Roberts RY, Kocherginsky M, et al (2009) Assessment of radiologist performance in the detection of lung nodules: dependence on the definition of "truth". Acad Radiol 16:28–38. \url{https://doi.org/10.1016/j.acra.2008.05.022}

\bibitem{Rubin2015}Rubin GD (2015) Lung nodule and cancer detection in computed tomography screening. J Thorac Imaging 30:130–138. \url{https://doi.org/10.1097/RTI.0000000000000140}

\bibitem{Aberle2013}
Aberle DR, DeMello S, Berg CD et al (2013) Results of the two incidence screenings in the national lung screening trial. N Engl J Med 369:920–931. \url{https://doi.org/10.1056/NEJMoa1208962}


\end{thebibliography}

\end{document}